\documentclass[sigconf]{acmart}

\usepackage{multirow}
\usepackage{colortbl}

\newcommand{\multitq}{\textsc{MultiTQ}}
\newcommand{\TimelineCronQR}{\mbox{\textsc{TimelineCronQ-R}}}

\AtBeginDocument{%
  }

\copyrightyear{2026}
\acmYear{2026}
\setcopyright{cc}
\setcctype{by}
\acmConference[CIKM '26]{Proceedings of the 35th ACM International Conference on Information and Knowledge Management}{November 07--11, 2026}{Rome, Italy}
\acmBooktitle{Proceedings of the 35th ACM International Conference on Information and Knowledge Management (CIKM '26), November 07--11, 2026, Rome, Italy}
\acmDOI{10.1145/3799682.3840716}
\acmISBN{979-8-4007-2539-5/2026/11}

\begin{document}

\title{SCoP: Structured Constraint Parsing for Evidence-Space Control in Temporal Knowledge Graph Question Answering}

\author{Xiaokun Guo}
\affiliation{%
  \institution{Institute of Information Engineering, Chinese Academy of Sciences}
  \city{Beijing}
  \country{China}
}
\affiliation{%
  \department{School of Cyber Security}
  \institution{University of Chinese Academy of Sciences}
  \city{Beijing}
  \country{China}
}
\email{guoxiaokun@iie.ac.cn}

\author{Zhen Xu}
\affiliation{%
  \institution{Institute of Information Engineering, Chinese Academy of Sciences}
  \city{Beijing}
  \country{China}
}
\affiliation{%
  \department{School of Cyber Security}
  \institution{University of Chinese Academy of Sciences}
  \city{Beijing}
  \country{China}
}
\email{xuzhen@iie.ac.cn}

\author{Dongdong Huo}
\authornote{Corresponding author.}
\affiliation{%
  \institution{Institute of Information Engineering, Chinese Academy of Sciences}
  \city{Beijing}
  \country{China}
}
\affiliation{%
  \department{School of Cyber Security}
  \institution{University of Chinese Academy of Sciences}
  \city{Beijing}
  \country{China}
}
\email{huodongdong@iie.ac.cn}

\author{Yanqiu Zhang}
\affiliation{%
  \institution{Institute of Information Engineering, Chinese Academy of Sciences}
  \city{Beijing}
  \country{China}
}
\affiliation{%
  \department{School of Cyber Security}
  \institution{University of Chinese Academy of Sciences}
  \city{Beijing}
  \country{China}
}
\email{zhangyanqiu@iie.ac.cn}

\author{Dongjin Yu}
\affiliation{%
  \institution{Institute of Information Engineering, Chinese Academy of Sciences}
  \city{Beijing}
  \country{China}
}
\affiliation{%
  \department{School of Cyber Security}
  \institution{University of Chinese Academy of Sciences}
  \city{Beijing}
  \country{China}
}
\email{yudongjin@iie.ac.cn}

\author{Yu Wang}
\affiliation{%
  \institution{Institute of Information Engineering, Chinese Academy of Sciences}
  \city{Beijing}
  \country{China}
}
\email{wangyu@iie.ac.cn}

\renewcommand{\shortauthors}{Guo et al.}

\begin{abstract}

Temporal Knowledge Graph Question Answering (TKGQA) requires answer inference from evidence that is both structurally valid and temporally admissible. Existing methods often leave anchor-event binding, temporal admissibility, and ordinal selection implicit in model reasoning, task-specific training, or similarity-driven retrieval, allowing locally relevant but invalid facts to enter the answer context. We formulate complex TKGQA as evidence-space control and propose \textbf{SCoP} (\textbf{S}tructured \textbf{Co}nstraint \textbf{P}arsing), a constraint-centric framework that externalizes temporal decisions before answer inference. Instead of treating retrieved facts as admissible evidence by default, SCoP separates answer-seeking event patterns from temporal anchor events, conservatively grounds them to canonical TKG entities and relations, and translates temporal intent into executable constraints with optional ranking requirements. These constraints operate over normalized point and interval ranges, enabling deterministic filtering of structurally compatible candidates and producing a compact evidence space for generation. Experiments on MultiTQ and TimelineCronQ-R assess SCoP across timestamped point-fact and interval-oriented settings with richer temporal relations and ordering dependencies. Without task-specific parameter updates, SCoP achieves 0.825 Hits@1 on MultiTQ and 0.761 Hits@1 on TimelineCronQ-R, with gains on constraint-intensive question types. These results support explicit evidence-space control over unconstrained retrieval or implicit temporal reasoning. 

\end{abstract}



\maketitle

\begin{figure}[!h]
    \centering
    \includegraphics[width=0.95\linewidth]{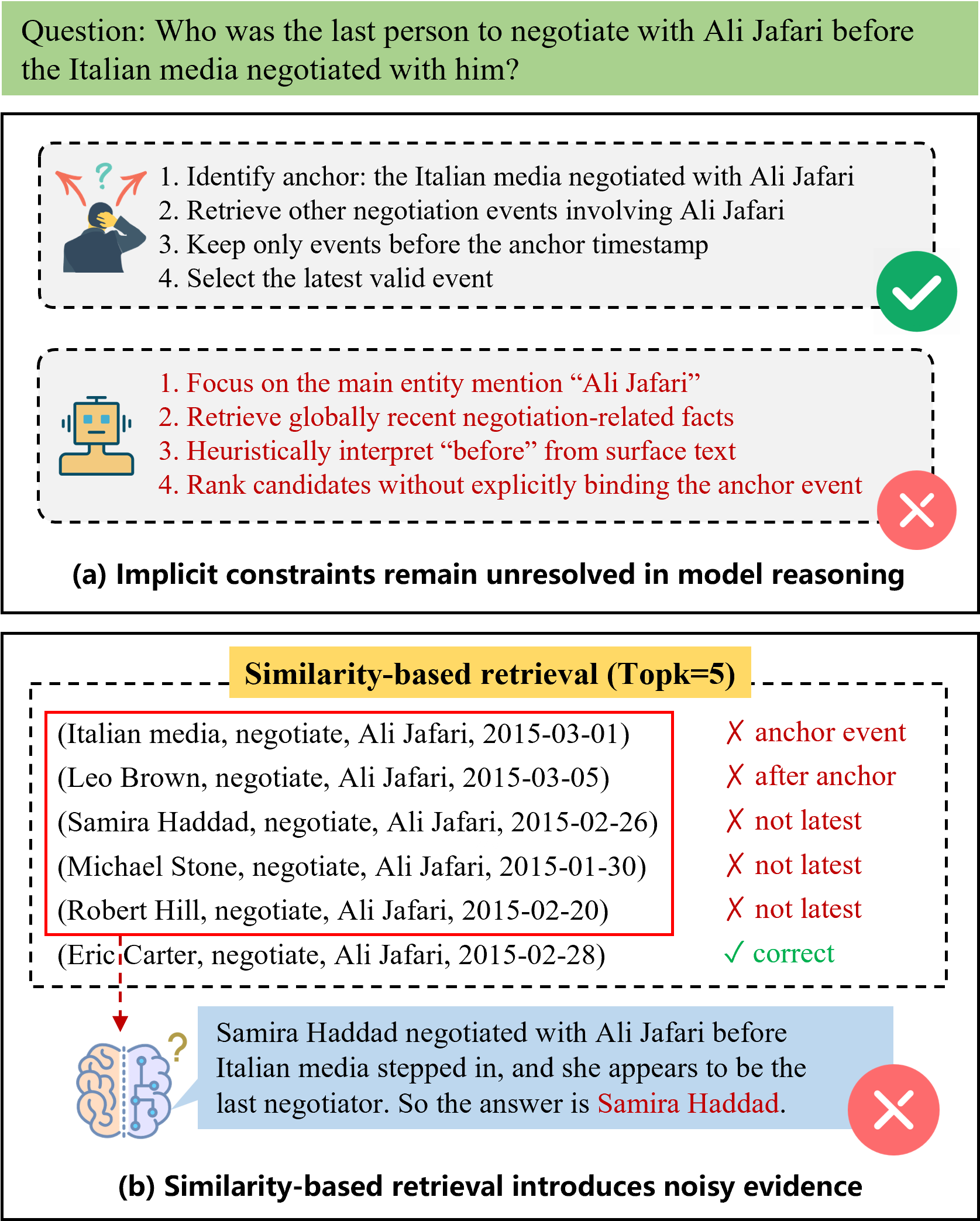}
    \caption{Two failure modes in TKGQA caused by implicit temporal constraints: model-internal decision errors and similarity-induced noisy evidence.}
    \label{fig:intro}
    \Description{Two-panel illustration using a temporal question about negotiations with Ali Jafari. The upper panel contrasts a constraint-aware reasoning path that identifies an anchor event, filters candidate events by the before relation, and selects the latest valid event with an unconstrained reasoning path that focuses on local entity relevance and returns an invalid result. The lower panel shows similarity-based top-k retrieval containing one valid candidate together with several candidates that occur after the anchor or fail the required ordering condition.}
\end{figure}

\section{Introduction}

Temporal Knowledge Graphs (TKGs) extend conventional knowledge graphs by attaching temporal annotations to facts, thereby enabling the representation of event order, temporal spans, and dynamic dependencies in evolving real-world knowledge~\cite{TrivediDWS17}.
Temporal Knowledge Graph Question Answering (TKGQA) aims to answer time-sensitive questions based on such structured evidence.
Unlike conventional KGQA, TKGQA requires a system to jointly satisfy structural conditions over entities and relations, as well as temporal conditions over timestamps, time intervals, event ordering, and temporal granularity.
Real-world temporal questions often involve multi-hop dependencies, dynamically changing factual states, and diverse temporal expressions such as \emph{before}, \emph{during}, and \emph{between}~\cite{jia2021complex, zhang2024mustq,liu2024towards}.
Therefore, a correct answer often depends on evidence that is admissible both structurally and temporally; even if a fact appears locally relevant, it may still lead to an incorrect conclusion if its structural or temporal conditions are not satisfied.
Although temporal annotations are already encoded in TKGs, selecting admissible evidence for complex temporal questions remains challenging.
Existing approaches may improve TKGQA through task-specific fine-tuning, model-generated reasoning traces, or similarity-driven retrieval, yet key temporal admissibility decisions are often still internalized in learned model behavior or approximated through retrieval relevance.

As shown in Fig.~\ref{fig:intro}(a), when anchor events, temporal conditions, temporal offsets, and ordering constraints remain implicit, the system may fail to bind them reliably to concrete event facts or execute the required temporal comparisons consistently.
As shown in Fig.~\ref{fig:intro}(b), similarity-based retrieval may return facts that are locally related to the question in terms of entities, relations, or timestamps, but fail to satisfy its complete structural or temporal requirements~\cite{gade2025abouttime}.
Such unconstrained top-$k$ contexts can enlarge the evidence space and increase the risk that the answer model reasons over inadmissible evidence~\cite{fayyaz2025collapse}.

We therefore study complex TKGQA from an evidence-space control perspective governed by question-induced constraints, rather than relying solely on model-internal temporal reasoning.
In this work, evidence-space control refers to constructing, before answer inference, a compact candidate set that is structurally compatible with the question and temporally admissible under its constraints.
To this end, we propose \textbf{SCoP} (\textbf{S}tructured \textbf{Co}nstraint \textbf{P}arsing), a constraint-centric TKGQA framework that externalizes key temporal decisions into structured executable constraints before answer generation.\footnote{\label{fn:repo}Code, prompts, and datasets are available at: \url{https://github.com/gxiaokun/scop}.}
Unlike full logical-form generation, SCoP does not specify a complete procedure for deriving the answer; instead, it structures the event and temporal conditions needed to determine which graph facts are admissible evidence.
Without task-specific parameter updates, SCoP identifies answer-seeking events and temporal anchor events, conservatively aligns them to canonical TKG elements, and applies executable temporal constraints to construct a compact, constraint-compliant evidence space for final answer inference.

We evaluate SCoP in two progressively more demanding TKGQA settings.
We first conduct experiments on \multitq{}~\cite{chen2023multi}, a standard multi-granularity benchmark that primarily evaluates question answering over timestamped point facts.
We then evaluate SCoP on \TimelineCronQR{}, a verified evaluation variant derived from the CronQuestions-KG subset generated by \textsc{TimelineKGQA}~\cite{sun2025timelinekgqa}.
In this benchmark, facts are associated with temporal intervals, and questions involve richer temporal relations, temporal operations, and ordering dependencies.
Results on both datasets show that SCoP remains consistently effective as temporal reasoning moves from point-based temporal facts to more complex interval-based and relation-intensive scenarios.
Overall, our main contributions are summarized as follows:






\begin{itemize}

\item \textbf{Structured constraint parsing for evidence-space control.}
We instantiate evidence-space control by externalizing question-induced structural and temporal admissibility conditions before answer inference. Rather than specifying a complete logical form for answer derivation, SCoP uses these structured conditions to determine which KG facts form the admissible evidence space for downstream answer inference.

\item \textbf{Structured decomposition of temporal questions into graph-compatible retrieval targets.}
SCoP separates answer-seeking event patterns from temporal anchor events and conservatively grounds the resulting structures to canonical TKG entities and relations, providing graph-compatible retrieval targets without introducing unsupported facts.
This separation keeps event retrieval structurally grounded while leaving temporal intent to the subsequent constraint layer.

\item \textbf{A structured, executable temporal constraint schema supporting deterministic filtering.}
SCoP represents event-referenced, explicit-time, and interval conditions as executable constraints over normalized point-or-interval ranges, and compiles them into deterministic admissibility predicates for candidate evidence. The schema supports conjunctive multi-anchor constraints, temporal granularity, and day-level offsets, while handling ordinal ranking separately from admissibility filtering.

\end{itemize}

\section{Related Work}

Earlier work on temporal and complex knowledge-base question answering explored semantic parsing and symbolic execution to explicitly represent temporal or compositional semantics through structured intermediate representations, including TEQUILA~\cite{jia2018tequila}, SYGMA~\cite{neelam2021sygma}, SF-TQA~\cite{ding2022semantic}, and Prog-TQA~\cite{chen2024self}.
These approaches instantiate structured reasoning in different forms: TEQUILA decomposes temporal questions into non-temporal subquestions and temporal constraints; SF-TQA uses a semantic framework of temporal constraints to guide query-graph generation; Prog-TQA generates symbolic program drafts that are aligned to the TKG and executed; and SYGMA produces a KB-agnostic logical representation with high-level reasoning constructs before KB-specific question mapping and answering.
SCoP is related to this line of work in making temporal reasoning conditions explicit, but uses its structured representation specifically to control which TKG facts form the admissible evidence space before downstream answer inference, rather than directly using it for query construction or answer execution.

A major line of TKGQA research encodes questions, entities, relations, and temporal annotations into continuous vector spaces, and predicts answers through learned scoring or similarity functions.
EmbedKGQA~\cite{saxena2020improving} uses KG embeddings for multi-hop KGQA, while CronKGQA~\cite{saxena2021question} and TempoQR~\cite{mavromatis2022tempoqr} further adapt representation-based reasoning to temporal QA over knowledge graphs by combining temporal KG embeddings with a question encoder.
MultiQA~\cite{chen2023multi} further studies multi-granularity temporal QA by modeling temporal semantics at different granularities.
Graph-enhanced variants, such as LGQA~\cite{liu2023local} and TwiRGCN~\cite{sharma2023twirgcn}, also incorporate temporal structure or temporally weighted message passing.
These methods improve temporal representation learning and candidate ranking, especially in sparse or multi-hop settings.
However, their evidence selection is still primarily governed by learned representations or similarity-based scoring.
For complex temporal questions, semantically related facts may still violate anchor-event conditions, interval boundaries, temporal granularity, offset constraints, or ordering requirements.

Recent work has also adapted LLMs, retrievers, or agent policies to temporal QA through task-specific training.
TimeR4~\cite{qian2024timer4} integrates rewriting, temporal retrieval, reranking, and generation for RAG-based TKGQA.
GenTKGQA~\cite{gao2024two} follows a two-stage framework of subgraph retrieval and answer generation, using LLM-guided temporal and structural cues together with graph signals.
PoK~\cite{qian2025pok} decomposes complex temporal questions into planned sub-objectives and retrieves temporally aligned facts from a temporal knowledge store.
Beyond TKGQA, Search-R1~\cite{jin2025search} shows that reinforcement learning can train LLMs to interact with external search environments during step-by-step reasoning.
More recently, Temp-R1~\cite{gong2026tempr1} formulates complex TKGQA as an autonomous agent problem and improves temporal reasoning through an expanded action space and reverse curriculum reinforcement learning.
These methods demonstrate that task-specific supervision, fine-tuned retrievers, instruction tuning, or reinforcement learning can improve temporal reasoning and tool-use behavior.
However, their effectiveness is often coupled with trained model parameters, supervised trajectories, or learned action policies, and temporal evidence selection is largely internalized into model behavior.

Another line of work uses LLMs as planners, decomposers, or reasoning controllers without necessarily updating model parameters.
ARI~\cite{chen2024temporal} induces abstract reasoning procedures from historical temporal QA examples and separates knowledge-agnostic reasoning guidance from knowledge-based answering.
TempAgent~\cite{hu2025tempagent} adapts the ReAct paradigm to TKGQA and introduces temporal-domain tools for interactive reasoning.
RTQA~\cite{gong2025rtqa} recursively decomposes complex temporal questions into sub-problems, solves them bottom-up, and aggregates multiple reasoning paths to mitigate error propagation.
MemoTime~\cite{tan2025memotime} augments temporal reasoning with memory of verified reasoning traces and operator-aware retrieval strategies.
These methods improve over direct LLM answering by introducing planning, decomposition, tool invocation, or memory-based reuse.
However, key temporal decisions often remain in natural-language reasoning traces, including anchor-event selection, temporal-condition verification, and ordering-based candidate selection.
When such decisions are not compiled into explicit predicates, they can be difficult to verify or execute consistently, and noisy retrieved evidence may still enter the final answering context.

Overall, SCoP differs from task-trained temporal reasoners and prompt-based LLM strategies by treating complex TKGQA as explicit evidence-space control. Rather than relying on learned policies or natural-language reasoning traces to resolve temporal admissibility, it compiles question-induced constraints into executable filtering conditions before answer inference.

\section{Method}

\begin{figure*}[t]
    \centering
    \includegraphics[width=0.98\linewidth]{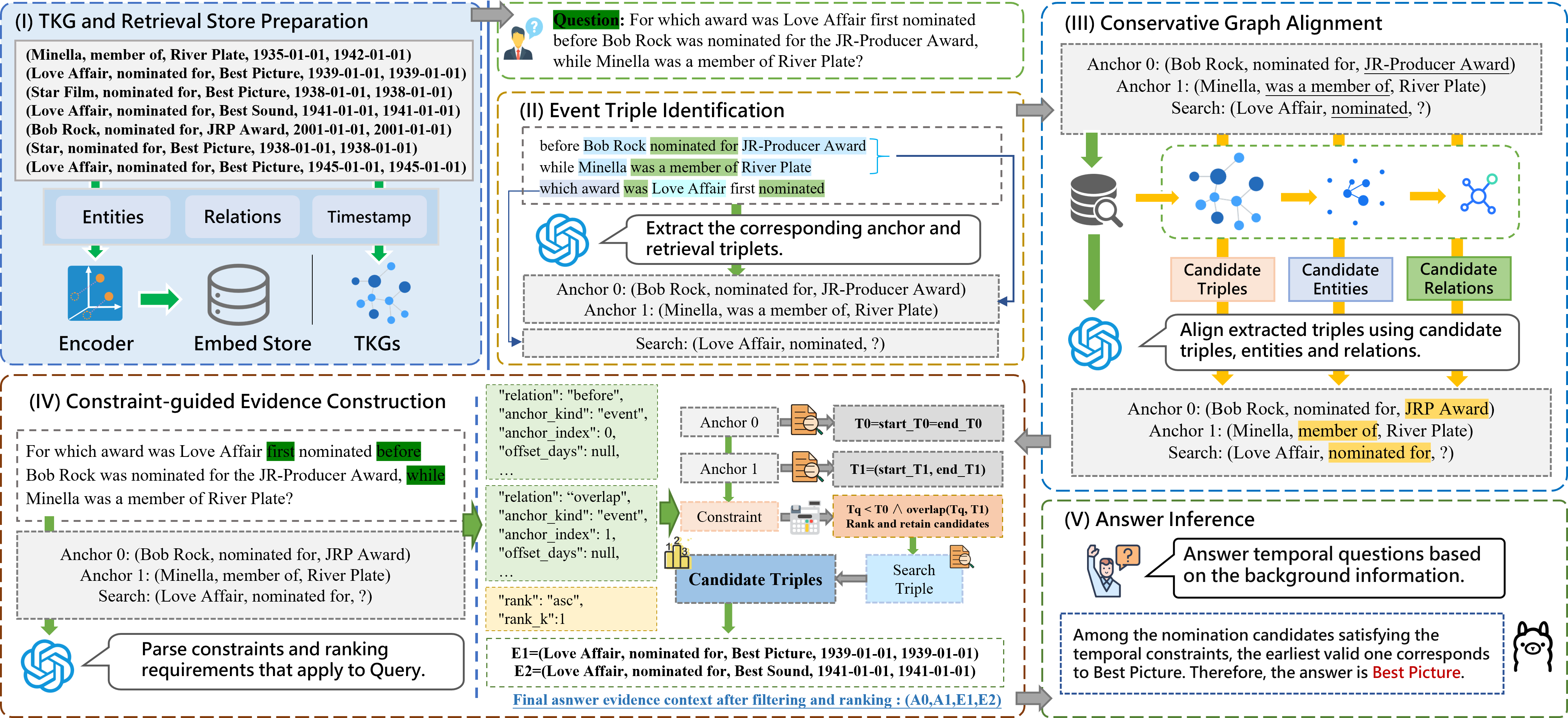}
    \caption{
    Overview of the SCoP framework. SCoP consists of one offline retrieval-store preparation stage, three query-time evidence-control stages, and a final answer inference stage: (I) Temporal Graph and Retrieval Store Preparation; (II) Event Triple Identification, which extracts anchor triples and an optional search triple; (III) Conservative Graph Alignment, which grounds extracted triples to canonical TKG elements; (IV) Constraint-guided Evidence Construction, which retrieves structurally compatible facts and filters them using parsed temporal constraints and ranking requirements; and (V) Answer Inference over the constrained evidence space.
}
    \label{fig:framework}
    \Description{Pipeline diagram of SCoP with five stages. An offline temporal knowledge graph and retrieval store provide entities, relations, and timestamps. At query time, a sample question is decomposed into two anchor triples and one search triple, which are conservatively aligned to canonical graph entities and relations. Temporal constraints and a ranking requirement are then parsed and applied to structurally compatible candidate triples. The filtered and ranked evidence is finally supplied to the answer inference stage.}
\end{figure*}

Given a temporal knowledge graph \(G\) composed of temporal facts \(x\), SCoP performs temporal question answering by first constructing a controlled evidence space and then conducting answer inference over the constrained evidence context. Specifically, SCoP consists of one offline retrieval-store preparation stage and three query-time evidence-control stages: event triple identification, conservative graph alignment, and constraint-guided evidence construction. The query-time stages first extract question-relevant event structures, then ground them to canonical TKG elements, and finally execute question-induced temporal constraints and optional ranking requirements over structurally compatible candidate facts. The resulting constrained evidence is passed to the answer generation module, so that final inference is performed over a temporally controlled evidence context rather than an unconstrained retrieval set. Figure~\ref{fig:framework} presents an overview of the SCoP framework.

\subsection{Temporal Graph and Retrieval Store Preparation}

Before question-specific reasoning, SCoP organizes the temporal
knowledge graph into a graph-backed retrieval store. We represent
it as a directed temporal graph
\[
G=(V,E), \qquad x=(s,r,o,t_s,t_e)\in E,
\]
where \(V\) is the entity set, and each \(x\) is a directed edge from
\(s\) to \(o\), labeled by its canonical relation \(r\) and temporal
scope \((t_s,t_e)\).

To uniformly support point events and interval-valued facts, SCoP preserves the temporal scope of each edge in a normalized point-or-interval form.
For a fact \(x\), we denote this stored temporal scope by \(\tau_x\), where \(t_s=t_e\) corresponds to a point event and \(t_s<t_e\) corresponds to an interval fact.
This representation is later used to resolve anchor-event times and execute temporal filtering.
To support schema grounding and candidate retrieval, SCoP builds dense indices over the canonical entity and relation vocabularies using an embedding encoder \(\phi(\cdot)\) and FAISS~\cite{johnson2021billion}.
At query time, these indices and the graph-backed retrieval store provide candidate entities, relations, and graph-compatible fact patterns for subsequent alignment and constrained evidence construction.
Structural validity and temporal admissibility are determined only in later stages through conservative graph alignment and explicit constraint execution.

\subsection{Event Triple Identification}

At this stage, SCoP transforms the question into retrieval-oriented structured event representations, and further maps it into an event pattern consisting of a set of known anchor events and at most one event to be retrieved:
\[
(\mathcal{A}_0(q),u_0(q))
=
\mathrm{Parse}_{\mathrm{evt}}(q),
\]
where
\[
\mathcal{A}_0(q)=[a_0,\ldots,a_{m-1}],
\qquad
a_i=(s_i,r_i,o_i),
\]
denotes the extracted anchor triples, and \(u_0(q)\) denotes the extracted search triple.

Each anchor triple is a complete event description and never contains the unknown slot ``\texttt{?}'', whereas the search triple, if present, is the only structure that may contain it.
For entity-seeking questions, SCoP constructs
\[
u_0(q)=(s,r,o),
\]
with exactly one unknown slot in either the subject or object position, while keeping the relation slot explicit.
For questions whose target is not an unknown entity, such as pure timestamp lookup or closed-target temporal-computation queries, SCoP sets
\[
u_0(q)=\varnothing,
\]
and retains the relevant complete events as anchor triples for downstream timestamp lookup, temporal comparison, or answer realization.
For example, given the question
\emph{``When was Bob Rock nominated for the JR-Producer Award?''},
SCoP retains
\[
a_0=(\text{Bob Rock},\text{nominated for},\text{JR-Producer Award})
\]
as an anchor triple and sets \(u_0(q)=\varnothing\).

A key design principle is that extracted triples encode only event content, not temporal intent.
Temporal expressions and operators, including \emph{before}, \emph{after}, explicit dates or intervals, \emph{same day/month/year}, and ordinal modifiers such as \emph{first} or \emph{last}, are excluded from triple slots and handled later as executable temporal constraints or ranking requirements.
This separation keeps the extracted triples suitable for graph retrieval and prevents temporal conditions from being conflated with entities or relations.

Anchor triples are introduced only when a complete event is explicitly stated in the question or can be minimally recovered from a local parallel reference.
Pure temporal expressions, such as \emph{before 2005}, \emph{after July 2014}, or \emph{in May 2009}, do not form anchors by themselves.
At this stage, relation phrases remain close to the question wording and are not required to match canonical TKG relation names; canonicalization is performed in the subsequent conservative graph alignment stage.


\subsection{Conservative Graph Alignment}

Triples extracted from natural language may use surface-form entity names or relation phrases that do not exactly match the canonical schema of the temporal knowledge graph.
SCoP therefore performs conservative graph alignment before constraint execution, normalizing extracted event structures into graph-compatible forms while preserving their original query semantics rather than rewriting the question or inferring missing facts.

Given \((\mathcal{A}_0(q),u_0(q))\), SCoP constructs a local candidate schema space
\[
\mathcal{V}_q=
(\mathcal{E}_q,\mathcal{R}_q,\mathcal{F}_q),
\]
where \(\mathcal{E}_q\), \(\mathcal{R}_q\), and \(\mathcal{F}_q\) denote retrieved candidate entities, relations, and fact triples, respectively.

For each extracted triple pattern, SCoP builds graph-backed candidate facts via slot-level retrieval.
Complete anchor triples retrieve candidates for all three slots; search triples retrieve candidates only for the observed slots and complete the unknown slot by enumerating compatible triples already present in \(G\), ensuring that all retained candidates correspond to valid graph facts.
For a candidate fact \(f=(s,r,o)\), its retrieval score is computed as
\[
\mathrm{Score}(f)
=
\mathrm{sim}(s,\tilde{s})
+
\mathrm{sim}(r,\tilde{r})
+
\mathrm{sim}(o,\tilde{o}),
\]
where \((\tilde{s},\tilde{r},\tilde{o})\) denotes the extracted surface-form triple.
For the unknown slot, SCoP sets \(\mathrm{sim}(z,\tilde{z})=1\) when \(\tilde{z}=\texttt{?}\).
This constant contribution does not affect the relative ranking among candidates from the same search pattern.
Candidate facts are then ranked by \(\mathrm{Score}(f)\) and retained in \(\mathcal{F}_q\) for downstream alignment.

The alignment module then outputs
\[
(\mathcal{A}(q),u(q))
=
\mathrm{Align}(q,\mathcal{A}_0(q),u_0(q),\mathcal{V}_q),
\]
where
\[
\mathcal{A}(q)=[\hat{a}_0,\ldots,\hat{a}_{m-1}]
\]
is the aligned anchor list, and \(u(q)\) is the aligned search triple or null.

SCoP adopts a conservative normalization strategy.
The alignment output preserves the number and order of anchor triples, represents every aligned triple as a three-field structure \((s,r,o)\), and never fills or removes the unknown slot ``\texttt{?}''.
By default, the original head--tail orientation and the position of ``\texttt{?}'' are preserved.
Only the search triple may undergo conservative orientation repair: conditioned on the original question and anchor-supported candidate evidence, the alignment model may reverse its extracted direction when the reversed orientation is judged to better preserve the intended retrieval semantics.
The original question is used only for semantic disambiguation, such as resolving local coreference, role mentions, or relation-action ambiguity, and is not used to introduce new entities, relations, or facts.
When no reliable normalization is available, the original surface form is retained.

Anchor and search triples are aligned differently.
Because anchor triples later serve as temporal references, SCoP prioritizes complete event-level grounding for each
\[
a_i=(s_i,r_i,o_i).
\]
When a reliable candidate fact in \(\mathcal{F}_q\) matches the complete event structure, SCoP adopts that fact-level grounded triple.
Otherwise, it applies conservative slot-level normalization over \(\mathcal{E}_q\) and \(\mathcal{R}_q\): each entity or relation slot is canonicalized only when its mapping is sufficiently reliable, while uncertain slots retain their original surface forms.
This design avoids fabricating fully grounded events by combining partial matches from unrelated graph facts.

Because the search triple is an open event pattern rather than a closed fact, SCoP normalizes only its known entity slots and relation phrase:
\[
u_0(q)=(s,r,o)
\quad\Longrightarrow\quad
u(q)=(\hat{s},\hat{r},\hat{o}),
\]
while keeping the unknown slot open.
Across both anchor and search alignment, SCoP enforces entity-type consistency and relation-action consistency, preventing surface similarity from altering the intended event semantics.

In the running example of Figure~\ref{fig:framework}, conservative alignment normalizes \emph{JR-Producer Award}, \emph{was a member of}, and \emph{nominated} to the graph-compatible forms \emph{JRP Award}, \emph{member of}, and \emph{nominated for}, respectively, while preserving the open answer slot.

\subsection{Constraint-guided Evidence Construction}

After event triple identification and conservative graph alignment, SCoP converts the question into graph-executable structural retrieval conditions.
If a search triple is present, the aligned triple defines an open but structurally restricted candidate space; otherwise, aligned complete event triples serve as closed retrieval targets for timestamp lookup or downstream temporal processing.
For search-triple questions, temporal conditions that cannot be encoded in retrieval triples themselves, such as explicit dates, event-relative ordering, intervals, and offsets, are parsed into executable constraints that remove structurally matched but temporally inadmissible facts.

Rather than assigning a single temporal-relation label to the whole question, SCoP parses an ordered constraint list together with an optional post-filter ranking object:
\[
(\mathcal{K}(q),\rho(q))
=
\mathrm{Parse}_{\mathrm{tmp}}(q,\mathcal{A}(q),u(q)),
\]
where
\[
\mathcal{K}(q)=[\kappa_0,\ldots,\kappa_n]
\]
denotes temporal admissibility constraints, and
\[
\rho(q)\in
\left(
\{\text{asc},\text{desc}\}\times\mathbb{N}_{>0}
\right)
\cup\{\varnothing\}
\]
denotes an optional ordinal ranking requirement over admissible candidates.

When a search triple is present, \(\mathrm{Parse}_{\mathrm{tmp}}\) instantiates temporal admissibility constraints and, when needed, ordinal ranking requirements over the open answer-bearing candidate space.
When \(u(q)=\varnothing\), the same parsing stage resolves the question into an anchor-only closed-target retrieval mode: the aligned complete event triples directly specify the facts to be retrieved for timestamp lookup or downstream temporal answer realization.
Since this mode introduces no open answer-bearing candidate space that requires additional temporal admissibility filtering, it yields
\(\mathcal{K}(q)=[]\) and \(\rho(q)=\varnothing\).

Each temporal constraint has the form
\[
\kappa_i=(\alpha_i,\omega_i,\eta_i,g_i,\delta_i),
\]
where \(\alpha_i\) specifies the reference type, \(\omega_i\) the temporal operator, \(\eta_i\) the reference content, \(g_i\) the comparison granularity, and \(\delta_i\) an optional day-level offset.
Specifically,
\[
\alpha_i\in
\{\text{event},\text{explicit\_time},\text{explicit\_interval}\},
\]
\[
\omega_i\in
\{\text{before},\text{after},\text{equal},\text{inside},\text{overlap}\}.
\]
The reference payload \(\eta_i\) depends on \(\alpha_i\):
\[
\eta_i =
\begin{cases}
j, & \alpha_i=\text{event},\quad 
j\in\{0,\ldots,|\mathcal{A}(q)|-1\}, \\
\tau, & \alpha_i=\text{explicit\_time}, \\
(\tau_s,\tau_e), & \alpha_i=\text{explicit\_interval}.
\end{cases}
\]
Here, \(j\) indexes an aligned anchor event in \(\mathcal{A}(q)\), \(\tau\) denotes an explicit temporal expression, and \((\tau_s,\tau_e)\) denotes an explicit temporal interval.
Granularity is used only for \(\text{equal}\), with
\[
g_i\in\{\text{day},\text{month},\text{year}\},
\]
and is null otherwise.
The offset field \(\delta_i\) is used for relative day-offset expressions such as ``\(N\) days before'' or ``\(N\) days after''; it stores a positive day count, with the temporal direction determined by \(\omega_i\). Ranking is separated from admissibility filtering: \((\text{asc},k)\) and \((\text{desc},k)\) indicate earlier- and later-oriented ordinal preferences, respectively, and retain a compact rank-focused subset to accommodate ties, temporal granularity ambiguity, and multiple valid answers. 
The operator inventory is designed as a compact set of executable primitives rather than an exhaustive taxonomy of temporal expressions.
More complex conditions are represented through their combination with reference types, granularity, offsets, and ranking requirements: before and after express directional comparisons; in and during correspond to interval inclusion; overlap captures temporal overlap; and between can be represented by an explicit interval or paired after–before constraints.
Ordinal modifiers such as first and last are handled separately through post-filter ranking.
Table~\ref{tab:constraint_schema} summarizes the structured temporal constraint and ranking schema used by SCoP.

\begin{table}[t]
\caption{Structured temporal constraint and ranking schema.}
\label{tab:constraint_schema}
\centering

\begin{tabular}{@{}c @{\hspace{4pt}} p{0.34\columnwidth} @{\hspace{5pt}} p{0.51\columnwidth}@{}}
\toprule
\multicolumn{1}{c}{\textbf{Field}}
& \multicolumn{1}{c}{\textbf{Values}}
& \multicolumn{1}{c}{\textbf{Meaning}} \\
\midrule

\(\alpha\)
& event / explicit\_time / explicit\_interval
& Type of temporal reference used to resolve the constraint. \\

\(\omega\)
& before / after / equal / inside / overlap
& Temporal operator for comparing candidate and reference times. \\

\(\eta\)
& anchor index / time text / interval pair
& Concrete reference content used during constraint execution. \\

\(g\)
& day / month / year / \(\varnothing\)
& Granularity for \text{equal}. \\

\(\delta\)
& positive days / \(\varnothing\)
& Day offset for relative time. \\

\(\rho\)
& \((\text{asc}, k)\), \((\text{desc}, k)\), or \(\varnothing\)
& Post-filter ordinal ranking. \\

\bottomrule
\end{tabular}
\end{table}

SCoP next defines the retrieval targets used to construct the initial candidate evidence space.
For entity-seeking questions, the aligned search triple \(u(q)\) is the target pattern; for time-seeking questions, \(u(q)=\varnothing\), and aligned complete event triples are used as closed target patterns for timestamp retrieval:
\[
\mathcal{U}(q)=
\begin{cases}
\{u(q)\}, & u(q)\neq \varnothing,\\
\mathcal{A}(q), & u(q)=\varnothing \land |\mathcal{A}(q)|>0.
\end{cases}
\]
Given \(\mathcal{U}(q)\), SCoP retrieves structurally compatible candidate facts:
\[
C(q)=\{x\in E \mid \exists u\in\mathcal{U}(q),\Gamma(x,u)\},
\]
where each temporal fact \(x\in E\) is represented as
\(x=(s_x,r_x,o_x,\tau_x)\).
For \(u=(s_u,r_u,o_u)\), structural compatibility is defined as
\[
\Gamma(x,u)=
\mathbb{I}[r_x\equiv r_u]\cdot
\mathbb{I}[s_u=\texttt{?} \vee s_x\equiv s_u]\cdot
\mathbb{I}[o_u=\texttt{?} \vee o_x\equiv o_u],
\]
where \(\equiv\) denotes schema-level equivalence after alignment.
Each parsed constraint \(\kappa_i\) is then compiled into an executable admissibility predicate:
\[
\Theta_{\kappa_i}(x;G)\in\{0,1\}.
\]
During execution, each candidate or reference timestamp is normalized into a closed date range
\[
\bar{\tau}(z)=[l_z,u_z].
\]
Day-level timestamps satisfy \(l_z=u_z\), while month- and year-level expressions are expanded to their corresponding calendar spans.
For event-referenced constraints, the indexed anchor triple may retrieve multiple graph facts; SCoP retains all resolvable anchor time ranges and regards a candidate as valid if it satisfies the constraint with respect to at least one such range.
If no resolvable anchor time is available, the corresponding constraint yields no admissible candidates.
Explicit time expressions and explicit intervals are normalized in the same form.

Given a candidate range \(X=[l_x,u_x]\) and a reference range \(A=[l_a,u_a]\), SCoP evaluates
\[
\text{before}(X,A):\ u_x<l_a, \quad
\text{inside}(X,A):\ l_x\ge l_a \land u_x\le u_a
\]
\[
\text{after}(X,A):\ l_x>u_a, \quad
\text{overlap}(X,A):\ l_x\le u_a \land u_x\ge l_a.
\]
For \(\text{equal}\), both ranges are first coarsened to the required granularity \(g_i\), and the predicate holds when the coarsened ranges overlap.
For offset constraints, \(\text{after}\) derives target dates from
\[
\{l_a+\delta_i,\;u_a+\delta_i\},
\]
while \(\text{before}\) derives target dates from
\[
\{l_a-\delta_i,\;u_a-\delta_i\}.
\]
A candidate satisfies the offset constraint if its normalized range covers at least one derived target date.
Conjunctive execution of all parsed constraints yields the filtered evidence space:
\[
\widetilde{C}(q)=
\{x\in C(q)\mid
\bigwedge_{\kappa_i\in\mathcal{K}(q)}
\Theta_{\kappa_i}(x;G)\}.
\]
This formulation naturally supports multi-anchor questions, where different constraints may refer to different temporal references and must hold simultaneously.

For the running example, the temporal parser produces
\[
\kappa_0=(\text{event},\text{before},0,\varnothing,\varnothing),
\qquad
\kappa_1=(\text{event},\text{overlap},1,\varnothing,\varnothing),
\]
together with the ranking requirement
\(\rho(q)=(\text{asc},1)\).
The two constraints jointly enforce the anchor-relative temporal
conditions, while the ranking object captures the ordinal modifier
\emph{first}.

After temporal filtering, SCoP applies the parsed ranking object:
\[
C^{*}(q)=
\begin{cases}
\mathrm{RankSelect}(\widetilde{C}(q);\rho(q)),
& \rho(q)\neq\varnothing,\\
\widetilde{C}(q),
& \rho(q)=\varnothing.
\end{cases}
\]
Here, $\mathrm{RankSelect}$ sorts admissible candidates by the start date of their normalized temporal ranges and retains rank-oriented subsets according to $\rho(q)$. To handle ties and temporal ambiguity, \emph{first} and \emph{last} preserve the top three facts, while explicit ordinal constraints (e.g., ``the fifth'') retain candidates up to the requested rank.

Finally, SCoP formats the retrieved temporal facts into a compact answer context \(\mathcal{H}(q)\), and invokes
\[
a=\mathrm{LLM}(q,\mathcal{H}(q)).
\]
For questions with a search triple, \(\mathcal{H}(q)\) contains the retrieved anchor evidence together with the filtered and ranked search facts.
For questions without a search triple, it contains the retrieved anchor facts used for downstream answer realization.
As a result, final inference is performed over a structurally controlled evidence context, with explicit temporal admissibility filtering applied whenever an open answer-bearing candidate space is present.

\section{Experiment}

\paragraph{Datasets.}
We evaluate SCoP on two TKGQA datasets: \multitq{}~\cite{chen2023multi} and \TimelineCronQR{}.
\multitq{} is a large-scale multi-granularity benchmark constructed from ICEWS05--15~\cite{garcia2018learning}, and we use it to evaluate SCoP under a standard temporal QA setting where facts are primarily represented by timestamps at different granularities.
\TimelineCronQR{} reconstructs the CronQuestions-KG subset generated by \textsc{TimelineKGQA}~\cite{sun2025timelinekgqa}, which is derived from CronQuestions~\cite{saxena2021question}.
Compared with \multitq{}, it is built on an interval-oriented temporal KG, where facts are associated with temporal scopes defined by start and end times.
In addition, \textsc{TimelineKGQA} characterizes temporal question complexity along four dimensions: context complexity, answer focus, temporal relations, and required temporal capabilities. In particular, its temporal relation space covers all 13 Allen interval relations, together with time-range set operations, duration-based operations, and temporal ranking. Therefore, \TimelineCronQR{} allows us to examine SCoP in a substantially richer interval-oriented setting, involving diverse interval relations, temporal arithmetic, and ordering dependencies.
Dataset statistics are summarized in Table~\ref{tab:dataset_stats}.

\paragraph{\TimelineCronQR{} Reconstruction.}
We found that the QA annotations of \textsc{TimelineKGQA}~\cite{sun2025timelinekgqa} contain duplicated or fragmented records, inconsistent answer representations, weakly grounded supporting events, and instances that cannot be reliably verified against the underlying temporal KG, which may compromise evaluation reliability.
We therefore conservatively reconstruct the QA annotation layer while keeping the temporal KG unchanged, using deterministic temporal execution, KG-backed completion where unambiguously resolvable, and consistency filtering to repair answer sets, remove unverifiable or invalid cases, resolve duplication, and normalize answer and event representations.
Surface-form rewriting is applied only when needed for naturalness and never changes answers, temporal relations, or supporting-event semantics.
Full reconstruction rules, prompts, verification procedures, and stage-wise statistics are provided in our repository.\textsuperscript{\ref{fn:repo}}

\begin{table}[ht]
\centering
\caption{Statistics of KGs and QA datasets. TCronQ-R denotes TimelineCronQ-R.}
\label{tab:dataset_stats}

\setlength{\tabcolsep}{1pt}
\begin{tabular*}{\columnwidth}{@{\extracolsep{\fill}}lccccccc@{}}
\toprule
\multirow{2}{*}{\textbf{Dataset}} & \multicolumn{4}{c}{\textbf{Knowledge Graph}} & \multicolumn{3}{c}{\textbf{Question-Answer}} \\
\cmidrule(lr){2-5}\cmidrule(lr){6-8}
& \textbf{Fact} & \textbf{Ent.} & \textbf{Rel.} & \textbf{Time.} & \textbf{Train} & \textbf{Val} & \textbf{Test} \\
\midrule
MultiTQ & 461,329 & 10,488 & 251 & 4,017 & 386,787 & 57,979 & 54,584 \\
TCronQ-R & 328,635 & 122,569 & 203 & 12,191 & 16,639 & 2,080 & 2,080 \\
\bottomrule
\end{tabular*}
\end{table}

\paragraph{Baselines and Settings.}
On \multitq{}, we compare SCoP with three groups of baselines: TKG embedding-based methods, including EmbedKGQA~\cite{saxena2020improving}, CronKGQA~\cite{saxena2021question}, and MultiQA~\cite{chen2023multi}; training-based LLM methods, including Search-R1~\cite{jin2025search}, TimeR4~\cite{qian2024timer4}, PoK~\cite{qian2025pok}, and Temp-R1~\cite{gong2026tempr1}; and strategy-based LLM methods, including ARI~\cite{chen2024temporal}, TempAgent~\cite{hu2025tempagent}, MemoTime~\cite{tan2025memotime}, and RTQA~\cite{gong2025rtqa}. 
Together, these baselines cover task-specific training, retrieval-augmented reasoning, agentic tool use, memory-augmented reasoning, and recursive question decomposition.
On \TimelineCronQR{}, where the complexity of its temporal relations makes straightforward adaptation infeasible for most existing methods, we employ a controlled evaluation setting for strategy-based retrieval methods.
We compare SCoP with three groups of baselines:
static single-turn retrieval methods, including RAG~\cite{lewis2020retrieval} and its filtering variant;
retrieval methods enhanced by query transformation, including HyDE~\cite{gao2023hyde} and Query2doc (Q2D)~\cite{wang2023query2doc};
and dynamic interleaved retrieval methods, including ReAct~\cite{yao2023react}, IRCoT~\cite{trivedi2023interleaving}, and their filtering variants.
All methods use the same final evidence budget of \(top\text{-}k=20\). Filtering variants first retrieve the top-50 candidates and then apply model-based filtering to retain 20 contexts. Iterative retrieval methods use at most 5 reasoning steps, with filtering applied at each step. SCoP is likewise restricted to a maximum retained evidence budget of 20.

For all dense retrieval components, we use BGE-M3~\cite{chen2024bge} as the unified embedding encoder, including the FAISS-based schema candidate retrieval in SCoP and the retrieval modules of the comparison methods on \TimelineCronQR{}.
For both \multitq{} and \TimelineCronQR{}, we use Qwen2.5-14B-Instruct as the backbone answer model with temperature set to 0. The SCoP results reported in Tables~\ref{tab:multi_results} and~\ref{tab:corn_r_results} are obtained on the full test set of each dataset. On \multitq{}, baseline results are collected from~\cite{gong2026tempr1}, whereas on \TimelineCronQR{}, all compared methods are evaluated under the same controlled setting.

\subsection{Main Results}

\paragraph{Performance comparison on MultiTQ}
Table~\ref{tab:multi_results} compares SCoP with existing methods on \multitq{}.
SCoP achieves the best overall Hits@1 score of 0.825, outperforming the strongest fine-tuning-based baseline Temp-R1 by 0.045 (+5.8\%) and the strongest strategy-based LLM baseline RTQA by 0.060 (+7.8\%).
Its advantage is most pronounced on the \emph{Multiple} question type, where SCoP reaches 0.653, exceeding Temp-R1 by 0.103 (+18.7\%) and RTQA by 0.229 (+54.0\%).
By contrast, on \emph{Single} questions, SCoP remains comparable to the strongest prior methods, indicating that its main gains arise in settings with richer temporal dependencies.
These results suggest that SCoP is particularly effective for constraint-intensive questions requiring the joint satisfaction of multiple structural and temporal conditions.
SCoP also maintains strong performance across answer types, achieving the best score on \emph{Entity} questions while remaining competitive on \emph{Time} questions.
Notably, these gains are obtained with Qwen2.5-14B-Instruct, whereas several previously reported strategy-based baselines rely on larger backbones, such as GPT-4-Turbo and DeepSeek-V3.

\begin{table}[htbp]
\caption{Performance comparison on the MultiTQ dataset. Results are reported as Hits@1. The best results are highlighted in bold, and the second-best results are underlined.}
\label{tab:multi_results}
\centering

\setlength{\tabcolsep}{3pt}

\begin{tabular*}{\columnwidth}{@{\hspace{3pt}}l@{\extracolsep{\fill}}ccccc@{\hspace{3pt}}}

\toprule
\rule{0pt}{3ex}
\multirow{2}{*}{\textbf{Method}}
& \multirow{2}{*}{\textbf{Overall}}
& \multicolumn{2}{c}{\textbf{Question Type}}
& \multicolumn{2}{c}{\textbf{Answer Type}} \\
\cmidrule(lr){3-4}\cmidrule(lr){5-6}
&
& \textbf{Single}
& \textbf{Multiple}
& \textbf{Entity}
& \textbf{Time} \\
\midrule

\rowcolor[gray]{0.92} \multicolumn{6}{c}{\textit{TKG Embedding-based Methods}} \\
EmbedKGQA & 0.206 & 0.235 & 0.134 & 0.290 & 0.001 \\
CronKGQA  & 0.279 & 0.337 & 0.134 & 0.328 & 0.156 \\
MultiQA   & 0.293 & 0.347 & 0.159 & 0.349 & 0.157 \\
\midrule

\rowcolor[gray]{0.92} \multicolumn{6}{c}{\textit{Fine-tuning-based LLM Methods}} \\
Search-R1$_{\scriptscriptstyle\text{Qwen2.5-7B}}$ & 0.352 & 0.474 & 0.094 & 0.230 & 0.705 \\
TimeR4$_{\scriptscriptstyle\text{GPT-3.5+LLaMA2}}$ & 0.728 & 0.887 & 0.335 & 0.639 & 0.945 \\
PoK$_{\scriptscriptstyle\text{GPT-4o+LLaMA2}}$ & 0.779 & \textbf{0.929} & 0.409 & 0.696 & 0.962 \\
Temp-R1$_{\scriptscriptstyle\text{LLaMA3.1-8B}}$ & \underline{0.780} & 0.888 & \underline{0.550} & \underline{0.714} & \textbf{0.969} \\
\midrule

\rowcolor[gray]{0.92} \multicolumn{6}{c}{\textit{Strategy-based LLM Methods}} \\
ARI$_{\scriptscriptstyle\text{GPT-3.5-Turbo}}$ & 0.380 & 0.680 & 0.210 & 0.394 & 0.344 \\
TempAgent$_{\scriptscriptstyle\text{GPT-4-Turbo}}$ & 0.702 & 0.857 & 0.316 & 0.624 & 0.870 \\
MemoTime$_{\scriptscriptstyle\text{DeepSeek-V3}}$ & 0.730 & 0.829 & 0.459 & 0.677 & 0.846 \\
RTQA$_{\scriptscriptstyle\text{DeepSeek-V3}}$ & 0.765 & \underline{0.902} & 0.424 & 0.692 & 0.942 \\
\midrule

\textbf{SCoP}$_{\scriptscriptstyle\text{Qwen2.5-14B-Instruct}}$
& \textbf{0.825}
& 0.895
& \textbf{0.653}
& \textbf{0.768}
& \underline{0.964} \\

\bottomrule
\end{tabular*}
\end{table}

\paragraph{Performance comparison on TimelineCronQ-R}
Table~\ref{tab:corn_r_results} summarizes the results on \TimelineCronQR{}.
SCoP achieves the best overall Hits@1 score of 0.761, outperforming the strongest \emph{dynamic interleaved retrieval} baseline, ReAct, by 0.130 (+20.6\%).
The gains are concentrated on the more challenging \emph{Medium} and \emph{Complex} questions.
Specifically, SCoP reaches 0.795 on \emph{Medium} questions, exceeding the strongest baseline RAG$_{\text{+Filter}}$ by 0.211 (+36.1\%), and achieves 0.673 on \emph{Complex} questions, surpassing ReAct$_{\text{+Filter}}$ by 0.173 (+34.6\%).
These results indicate that SCoP is particularly effective in settings requiring the joint enforcement of multiple temporal conditions.
On \emph{Simple} questions, SCoP remains competitive but trails direct retrieval methods, suggesting that explicit evidence-space control provides smaller marginal benefits when temporal constraint resolution is less demanding.

\begin{table}[htbp]
\caption{Performance comparison on TimelineCronQ-R. Results are reported as Hits@1. Baseline methods are evaluated both in their standard form and with a post-retrieval filter ($_{\text{+Filter}}$) to analyze the impact of noise reduction. The best results are highlighted in bold, and the second-best results are underlined.}
\label{tab:corn_r_results}
\centering


\setlength{\tabcolsep}{3.5pt}

\begin{tabular*}{\columnwidth}{@{\hspace{3pt}}l@{\extracolsep{\fill}}cccc@{\hspace{3pt}}}
\toprule
\rule{0pt}{3ex} \multirow{2}{*}{\textbf{Method}} 
& \multirow{2}{*}{\textbf{Overall}} 
& \multicolumn{3}{c}{\textbf{Question Level}} \\
\cmidrule(lr){3-5}
& 
& \textbf{Simple} 
& \textbf{Medium} 
& \textbf{Complex} \\

\midrule

\rowcolor[gray]{0.92} \multicolumn{5}{c}{\textit{Static Single-turn Retrieval}} \\
RAG                        & 0.515 & \textbf{0.840} & 0.498 & 0.209 \\
RAG$_{\text{+Filter}}$     & 0.575 & \textbf{0.840} & \underline{0.584} & 0.300 \\

\midrule
\rowcolor[gray]{0.92} \multicolumn{5}{c}{\textit{Query Transformation Enhancement}} \\
HyDE                       & 0.494 & 0.801 & 0.485 & 0.197 \\
HyDE$_{\text{+Filter}}$    & 0.439 & 0.727 & 0.341 & 0.249     \\
Q2D                        & 0.523 & 0.807 & 0.524 & 0.239 \\
Q2D$_{\text{+Filter}}$     & 0.438 & 0.683 & 0.384 & 0.246 \\

\midrule
\rowcolor[gray]{0.92} \multicolumn{5}{c}{\textit{Dynamic Interleaved Retrieval}} \\
ReAct                      & \underline{0.631} & \underline{0.820} & 0.576 & 0.499 \\
ReAct$_{\text{+Filter}}$   & 0.611 & 0.811 & 0.522 & \underline{0.500} \\
IRCoT                      & 0.577 & 0.807 & 0.440 & 0.486 \\
IRCoT$_{\text{+Filter}}$   & 0.573 & 0.795 & 0.432 & 0.493 \\

\midrule
\textbf{SCoP}              & \textbf{0.761} & 0.815 & \textbf{0.795} & \textbf{0.673} \\

\bottomrule
\end{tabular*}
\end{table}

Across both datasets, SCoP's strongest gains consistently appear in settings with higher temporal-constraint complexity: \emph{Multiple} questions on \multitq{} and \emph{Medium}/\emph{Complex} questions on \TimelineCronQR{}.
This pattern supports our central claim that complex TKGQA benefits from explicitly controlling structurally and temporally admissible evidence before answer inference.

\subsection{Influence of Backbone Models}

\begin{figure}[ht]
    \centering
    \includegraphics[width=1\linewidth]{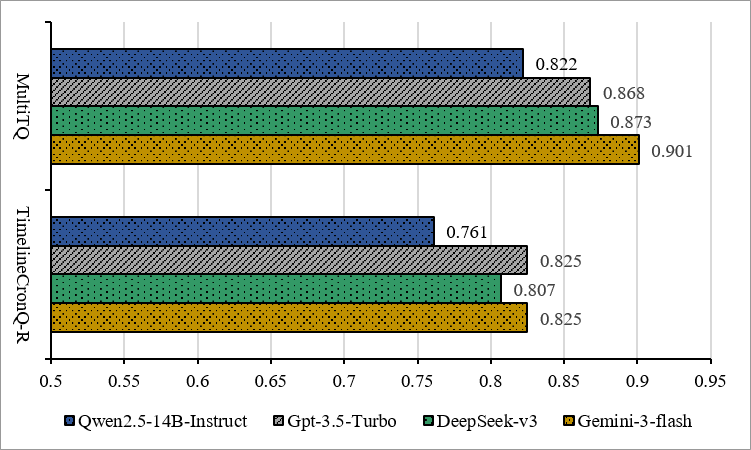} 
    \caption{Effect of different backbone models on the overall Hits@1 performance of SCoP.}
    \label{fig:backbone}
    \Description{Horizontal bar chart comparing SCoP with four backbone models on MultiTQ and TimelineCronQ-R. All three alternative backbones improve over Qwen2.5-14B-Instruct in overall Hits@1. Gemini-3-Flash achieves the highest result on MultiTQ, while GPT-3.5-Turbo and Gemini-3-Flash achieve the highest overall results on TimelineCronQ-R.}
\end{figure}

To examine the sensitivity of SCoP to backbone choice and facilitate a more direct comparison with representative strategy-based methods under comparable backbone configurations, we evaluate SCoP with Qwen2.5-14B-Instruct, Gemini-3-Flash, as well as GPT-3.5-Turbo and DeepSeek-V3, which are adopted by representative strategy-based baselines. We conduct experiments on 5,000 randomly sampled test instances from \multitq{} while preserving its original data distribution, and additionally evaluate on the full test set of \TimelineCronQR{}.
Figure~\ref{fig:backbone} and Table~\ref{tab:model_fine_grained_comparison} show that all alternative backbones achieve higher overall Hits@1 than the Qwen2.5-14B-Instruct setting, although the magnitude of improvement varies across question types.
On \multitq{}, the gains are substantially larger for \emph{Multiple} than for \emph{Single} questions; on \TimelineCronQR{}, improvements are generally more pronounced on \emph{Medium} and \emph{Complex} questions than on \emph{Simple} ones.

\begin{table}[htbp]
\centering
\caption{Detailed performance comparison (Hits@1) across distinct question complexity levels.}
\label{tab:model_fine_grained_comparison}

\begin{tabular*}{\columnwidth}{@{\hspace{3pt}} l @{\extracolsep{\fill}} c c c c c @{\hspace{3pt}}}
\toprule
\rule{0pt}{3ex}
\multirow{2}{*}{\textbf{Model}} 
& \multicolumn{2}{c}{\textbf{MultiTQ}} 
& \multicolumn{3}{c}{\textbf{TimelineCronQ-R}} \\
\cmidrule(lr){2-3}\cmidrule(lr){4-6}
& \textbf{Sing.} & \textbf{Mult.} 
& \textbf{Simp.} & \textbf{Med.} & \textbf{Comp.} \\
\midrule
Qwen2.5-14B-Inst   & 0.892 & 0.648 & 0.815 & 0.795 & 0.673 \\
GPT-3.5-Turbo  & 0.903 & \textbf{0.822} & 0.825 & \textbf{0.892} & \textbf{0.759} \\
DeepSeek-V3    & \underline{0.916} & 0.767 & \underline{0.844} & \underline{0.874} & 0.702 \\
Gemini-3-Flash & \textbf{0.952} & \underline{0.778} & \textbf{0.860} & 0.854 & \textbf{0.759} \\
\bottomrule
\end{tabular*}
\end{table}

\begin{table*}[!t]
\centering
\caption{Candidate-space control across question types on MultiTQ and TimelineCronQ-R.}
\begin{tabular*}{1\textwidth}{@{\hspace{3pt}} l @{\extracolsep{\fill}} c c c c c l c c c c @{\hspace{3pt}}}
\toprule

\multicolumn{5}{c}{\textbf{\textit{MultiTQ}}} & & \multicolumn{5}{c}{\textbf{\textit{TimelineCronQ-R}}} \\
\cmidrule(r){1-5} \cmidrule(l){7-11}

\textbf{Type} & \textbf{Count} & \textbf{Init. Evid.} & \textbf{Final Evid.} & \textbf{ECR(\%)} & &
\textbf{Type} & \textbf{Count} & \textbf{Init. Evid.} & \textbf{Final Evid.} & \textbf{ECR(\%)} \\
\midrule

equal         & 17212 & 747.00 & 3.58 & 99.52\% & & interval\_relation     & 21   & 16.57 & 1.52 & 90.81\% \\
equal\_multi   & 3207  & 519.98 & 5.39 & 98.96\% & & quantitative\_temporal & 234  & 60.07 & 2.00 & 96.67\% \\
before\_after  & 11031 & 146.37 & 9.93 & 93.22\% & & relative\_temporal     & 155  & 52.23 & 2.54 & 95.13\% \\
before\_last   & 6112  & 570.93 & 4.41 & 99.23\% & & temporal\_operation    & 455  & 8.60  & 5.81 & 32.45\% \\
after\_first   & 6262  & 633.19 & 3.87 & 99.39\% & & temporal\_order        & 336  & 45.26 & 7.46 & 83.53\% \\
first\_last    & 10425 & 13.02  & 2.36 & 81.91\% & & timeline\_structure    & 855  & 33.21 & 3.95 & 88.10\% \\
\midrule

Overall       & 54249 & 437.42 & 4.87 & 98.89\% & & Overall               & 2056 & 34.05 & 4.58 & 86.54\% \\

\bottomrule
\end{tabular*}
\label{tab:evidence_reduction_qtype}
\end{table*}

This pattern is consistent with the main results, suggesting that backbone capability becomes more influential as event parsing, grounding, and temporal constraint interpretation become more demanding.
At the same time, no single backbone dominates across all question categories, indicating complementary strengths across different temporal reasoning regimes.
These results also clarify the role of the backbone in SCoP: structured constraints externalize temporal admissibility for deterministic evidence filtering, while the quality of the parsed events, anchors, and constraints still depends on the language model.
Thus, SCoP is not tied to a specific backbone, while its absolute performance still depends on backbone capability.

\subsection{Evidence-Space Control Analysis}

To examine whether SCoP effectively controls the candidate evidence space, we compare the initial candidate evidence before constraint execution with the final evidence passed to the answer model after constraint-guided evidence construction. 
The quantified evaluation results are summarized in Table~\ref{tab:evidence_reduction_qtype}. Specifically, \textit{Initial Evid.} and \textit{Final Evid.} report the average evidence counts per evaluable sample within each temporal reasoning type.
For questions with a search triple, \textit{Initial Evid.} denotes the number of answer-bearing search-fact candidates before temporal constraint execution.
\textit{Final Evid.} denotes the number of retained search facts after executable constraint filtering, optional ranking, and final context truncation.
For questions without a search triple, \textit{Initial Evid.} denotes the summed candidate count retrieved for the aligned complete anchor triples.
\textit{Final Evid.} denotes the summed number of anchor facts retained in the bounded closed-target answer context.
We quantify the overall compression effect within each subset using the \textit{Evidence Compression Rate}:
\[
\mathrm{ECR}
=
1-
\frac{\sum_i \textit{Final Evid.}_i}
{\sum_i \textit{Initial Evid.}_i},
\]
where $i$ ranges over all evaluable samples in each subset.
After excluding samples with incomplete retrieval statistics, the evaluation covers 99.4\% of \multitq{} and 98.8\% of \TimelineCronQR{}.
The type-wise results further show that compression varies with temporal selectivity.
In \multitq{}, equality-, ranking-, and before/after-related questions are strongly compressed, whereas first-last questions exhibit lower reduction because their initial candidate spaces are already smaller.
In \TimelineCronQR{}, quantitative- and relative-temporal questions show the strongest compression, while temporal-operation questions retain more evidence for downstream answer realization.
These differences are consistent with the selectivity of the executable temporal constraints: tighter temporal conditions eliminate more structurally compatible candidates, whereas less selective conditions preserve a broader evidence context.

\begin{table}[htbp]
\centering
\caption{Gold-event and gold-answer retention in the final evidence space on \TimelineCronQR{}.}

\begin{tabular*}{\columnwidth}{@{\hspace{3pt}}l@{\extracolsep{\fill}}cccc@{\hspace{3pt}}}
\toprule
\rule{0pt}{3ex}
\multirow{2}{*}{\textbf{Level}} 
& \multicolumn{2}{c}{\textbf{Gold Event}} 
& \multicolumn{2}{c}{\textbf{Gold Answer}} \\
\cmidrule(lr){2-3} \cmidrule(lr){4-5}
& \textbf{Any Ret.} 
& \textbf{Full Ret.} 
& \textbf{Any Ret.} 
& \textbf{Full Ret.} \\
\midrule
Simple  & 84.3\% & 84.0\% & 87.1\% & 86.6\% \\
Medium  & 90.2\% & 84.4\% & 89.6\% & 82.2\% \\
Complex & 82.6\% & 76.4\% & 74.7\% & 72.4\% \\
\midrule
Overall & 85.7\% & 81.6\% & 84.9\% & 81.3\% \\
\bottomrule
\end{tabular*}
\label{tab:gold_evidence_preservation}
\end{table}

The compression analysis raises a further question: does the reduced evidence space still preserve the contextual support required for answer inference?
Because \multitq{} lacks gold event annotations for direct event-level retention analysis, we conduct this evaluation on \TimelineCronQR{} across different question difficulty levels.
We report retention at both the gold-event and gold-answer levels using two metrics:
\((1)\) \textit{Any Retention Rate}, the proportion of evaluable samples whose final evidence context contains at least one gold event or gold answer;
and \((2)\) \textit{Full Retention Rate}, the proportion preserving all gold events or gold answers.

As shown in Table~\ref{tab:gold_evidence_preservation}, SCoP retains a large majority of gold events and gold answers after compression, with overall Any/Full Retention rates of 85.7\%/81.6\% and 84.9\%/81.3\%, respectively.
Full Retention generally decreases with question complexity, with \emph{Complex} questions showing the lowest retention for both gold events and gold answers.
The gap between Any and Full Retention indicates that SCoP may preserve partial support while losing complementary evidence required jointly for answer inference, particularly on \emph{Complex} questions involving multiple temporally related facts.
Taken together, these results show that SCoP substantially contracts the evidence space while preserving most task-relevant support, although complete evidence preservation remains more difficult for compositionally demanding questions.


\subsection{Ablation Studies}

Having shown that SCoP compresses the candidate evidence space while preserving most task-relevant support, we further examine the contribution of its three core components on both \multitq{} and \TimelineCronQR{}.
All variants use Qwen2.5-14B-Instruct as the unified backbone, and the results are summarized in Table~\ref{tab:ablation_overall}.
We consider three variants.
For \emph{w/o triple}, we remove structured event-triple identification and replace the explicit answer-anchor event separation with a retrieval-based pseudo-decomposition strategy, where the original question is directly used to retrieve pseudo anchor triples and construct a pseudo search triple.
For \emph{w/o alignment}, we remove conservative graph alignment and directly use the raw extracted triples for downstream retrieval and constraint execution.
For \emph{w/o constraint}, we retain event-triple identification and graph alignment but disable temporal constraint parsing and execution, so that structurally compatible but temporally unfiltered candidate facts are passed to the answer model.
Together, these variants isolate whether SCoP's gains arise from explicit event-role decomposition, KG-compatible grounding, and executable temporal admissibility control.

\begin{table}[htbp]
\centering
\caption{Ablation study of SCoP under the Overall Hits@1 metric. \(\Delta\) denotes the relative performance change compared with the full model.}
\label{tab:ablation_overall}

\begin{tabular*}{\columnwidth}{@{\hspace{3pt}} l @{\extracolsep{\fill}} c c c c @{\hspace{3pt}}}
\toprule
\rule{0pt}{3ex}
\multirow{2}{*}{\textbf{Model Variant}} 
& \multicolumn{2}{c}{\textbf{MultiTQ}} 
& \multicolumn{2}{c}{\textbf{TimelineCronQ-R}} \\
\cmidrule(lr){2-3}\cmidrule(lr){4-5}
& \textbf{Hits@1} & \textbf{$\Delta$} 
& \textbf{Hits@1} & \textbf{$\Delta$} \\
\midrule
Full           & \textbf{0.825} & --      & \textbf{0.761} & --      \\
w/o triple     & 0.517 & -37.3\% & 0.420 & -44.8\% \\
w/o alignment  & 0.463 & -43.9\% & 0.721 & -5.3\%  \\
w/o constraint & 0.449 & -45.6\% & 0.541 & -28.9\% \\
\bottomrule
\end{tabular*}
\end{table}

As shown in Table~\ref{tab:ablation_overall}, all ablated variants underperform the full model on both datasets, showing that the three components contribute to different stages of evidence-space control. 
On \multitq{}, removing constraint execution causes the largest drop, reducing Hits@1 from 0.825 to 0.449, while removing graph alignment leads to a comparable decline to 0.463. 
This indicates that \multitq{} relies heavily on both temporal admissibility filtering and KG-compatible schema grounding: without constraints, structurally relevant but temporally invalid facts enter the answer context; without alignment, extracted event structures cannot be reliably matched to canonical entities and relations. 
On \TimelineCronQR{}, removing event-triple identification causes the largest degradation, from 0.761 to 0.420, suggesting that explicitly separating answer-seeking events from temporal anchors is particularly important in interval-oriented and relation-intensive questions. 
Disabling constraint execution also produces a substantial drop to 0.541, confirming that executable temporal constraints remain necessary for filtering admissible evidence. 
The smaller impact of removing alignment on \TimelineCronQR{} may reflect that its entity and relation mentions are often closer to KG-compatible forms after extraction. 
Overall, the ablation results support the coordinated design of SCoP: event-triple identification defines the answer and anchor evidence targets, graph alignment grounds them to the TKG schema, and constraint execution controls which structurally retrieved facts are temporally admissible for answer inference.

\section{Conclusion}

We present SCoP, a structured constraint parsing framework that controls the admissible evidence space for temporal knowledge graph question answering before answer inference.
By separating event identification, conservative graph alignment, and executable temporal constraint filtering, SCoP consistently improves performance on \multitq{} and \TimelineCronQR{}, with its strongest gains on constraint-intensive questions.
Further analysis shows that SCoP substantially reduces candidate evidence while preserving most task-relevant support, supporting explicit evidence-space control over unconstrained retrieval or implicit temporal reasoning.
These results validate the effectiveness of explicit evidence-space control, while also revealing several limitations of the current framework.
SCoP relies on off-the-shelf language models for event identification, constraint parsing, and graph alignment, so errors in these stages may propagate to downstream evidence construction.
Its current constraint schema does not yet cover all ambiguous or compositional temporal phenomena, and complete evidence preservation remains more challenging for complex questions requiring multiple supporting facts to survive jointly.
Future work will therefore focus on extending the constraint schema, improving multi-evidence preservation, and evaluating SCoP across broader temporal knowledge graphs and reasoning settings.

\section*{GenAI Usage Disclosure}
ChatGPT (GPT-5.5) was used for language polishing, translation, and assisting with refining and optimizing code written by the authors. Additionally, Gemini-3-Flash was used during the reconstruction of the \TimelineCronQR{} dataset for constrained surface-form rewriting of benchmark questions. Neither tool was involved in the design of the core algorithms or research methodology, which were developed by the authors. All AI-assisted code, data, and manuscript content were reviewed and verified by the authors, who take full responsibility for the final research outputs.

\bibliographystyle{ACM-Reference-Format}
\bibliography{reference}

\end{document}